# Robust Person Identification: A WiFi Vision-based Approach

Yili Ren
*Florida State University*

Yichao Wang
*Florida State University*

Sheng Tan
*Trinity University*

Yingying Chen
*Rutgers University*

Jie Yang
*Florida State University*

## Abstract

Person re-identification (Re-ID) has become increasingly important as it supports a wide range of security applications. Traditional person Re-ID mainly relies on optical camera-based systems, which incur several limitations due to the changes in the appearance of people, occlusions, and human poses. In this work, we propose a WiFi vision-based system, 3D-ID, for person Re-ID in 3D space. Our system leverages the advances of WiFi and deep learning to help WiFi devices "see", identify, and recognize people. In particular, we leverage multiple antennas on next-generation WiFi devices and 2D AoA estimation of the signal reflections to enable WiFi to visualize a person in the physical environment. We then leverage deep learning to digitize the visualization of the person into 3D body representation and extract both the static body shape and dynamic walking patterns for person Re-ID. Our evaluation results under various indoor environments show that the 3D-ID system achieves an overall rank-1 accuracy of 85.3%. Results also show that our system is resistant to various attacks. The proposed 3D-ID is thus very promising as it could augment or complement camera-based systems.

## 1 Introduction

Intelligent surveillance is gaining increasing attention due to the growing demand for public and private security [69]. It has been progressively deployed in public, workplace, and home, providing a wide range of security applications, such as facility security, perimeter monitoring, and abnormal behavior detection [26, 58]. Among these applications, person re-identification (Re-ID) is a fundamental one, which identifies individuals in public and private spaces across time and locations [68]. Person Re-ID is also a basic component for access control, suspicious behavior tracking, and physical intrusion detection [45]. The rising of the Internet of Things (IoT) is also boosting the adaptation of the person Re-ID system in many emerging applications [21]. For instance, an integrated person Re-ID system can facilitate customized services and prevent unauthorized access to security-sensitive resources in a smart home environment [80].

Traditional methods for person Re-ID mainly rely on computer vision approaches, in which the camera-captured optical images or videos are analyzed for human behavior monitoring and person Re-ID [22, 42]. Despite much progress has been made by incorporating deep learning, the camera-based systems have several limitations [5]. For example, it cannot work in non-line-of-sight (NLoS) or poor lighting conditions, such as when a person is behind an obstacle or in a dark environment [15]. Moreover, the optical camera-based system is sensitive to the variation in an individual's appearance due to changes in clothes, viewpoints, and unconstrained human poses [2]. It has to ensure that surveillance cameras do not point to private areas and the captured camera data is stored or transmitted securely [11, 14].

In recent years, WiFi devices have become more and more ubiquitous in public and private places including airports, train stations, offices, and homes. While nearly 1 billion surveillance cameras have been deployed worldwide in 2021 [9], this number is about 22.2 billion for WiFi devices [3]. The number of WiFi devices is thus an order of magnitude larger than that of the cameras. This difference will be further magnified by the increasing usage of IoT devices. For instance, the WiFi devices will increase by 50% in the next 4 years and will reach 31 billion by 2025 [63]. As both visible lights and WiFi signals are electromagnetic waves, we could leverage the more pervasive WiFi signals to illuminate the human body and analyze the reflections for person Re-ID to augment or complement camera-based systems. Compared to camera-based systems, the WiFi-based approach has several advantages. For example, it can work in NLoS or poor lighting conditions as the WiFi signal can traverse occlusions and illuminate the human body in dark environments. Moreover, as the WiFi signal traverses clothes but is reflected off the human body, it is less affected by an individual's appearances such as changes in clothes. In addition, the WiFi-based approach will not create sensitive data such as colorful or high-resolution visual data like that of the cameras.

Unlike cameras, the commodity WiFi devices do not have CCD/CMOS sensors to sense the intensity of the signal reflections at megapixel resolution. Instead, they rely on omnidirectional antennas to sense the total intensity of the signal reflected from the entire environment. Thus, it cannot provide any spatial information to distinguish objects or persons located at different physical locations. Existing work in using WiFi for human sensing uses a black-box approach by directly inputting the received WiFi signals into deep learning models for activity and person identification [44, 71, 72]. Their assumption is that similar activities or people will interrupt the WiFi signals similarly, resulting in similar signal change patterns. This may be useful for applications with pre-defined activities and controlled users but is less applicable for security applications, such as person Re-ID. It is because, given a signal change pattern, it may correspond to a large number of uncontrolled users or many unknown free-from activities. It is thus less robust for unseen environments and people. Moreover, the black-box-based deep learning approach is more sensitive to adversarial attacks, where maliciously interrupted WiFi signals can bypass the surveillance and person Re-ID system [28, 34, 73]. Indeed, recent work has demonstrated such a vulnerability in existing WiFi-based user identification systems [38].

In this work, we propose a WiFi vision-based approach for person Re-ID in 3D physical space. We leverage the advancement of WiFi technology and deep learning to help WiFi devices "see", identify, and recognize persons as we humans do. Our system, 3D-ID, helps WiFi devices "see" a person by leveraging (i) the multiple antennas on next-generation WiFi devices, and (ii) the two-dimensional angle of arrival (2D AoA) estimation of the WiFi signal reflections. First, the next generation of WiFi supports a fairly large number of antennas, which can be re-used to discern the signal reflected from spatially separated objects and persons, providing spatial resolution similar to that of the optical image. In particular, the new generation of WiFi 6 devices support up to 8 antennas [25], whereas the next generational WiFi 7 further increases it to 16 antennas [12]. With spatially distributed antennas at the WiFi receiver, the signal reflections from the different directions could be separated with signal processing techniques, providing the theoretical foundation to derive spatial information of the physical space. Second, we leverage the 2D AoA of the signal reflections to visualize a person in the physical space. In particular, we derive the 2D AoA in terms of the azimuth and elevation, where azimuth is an angular measurement of the signal reflection on the horizon and elevation is the angular measurement of the same reflection in the vertical direction. With multiple antennas and the 2D AoA estimation, the WiFi devices could generate a visualization of the signal reflections from the surrounding environment, thus providing the ability for the WiFi devices to "see" a person in the physical world as we humans do.

We then propose to extract intrinsic features of a person including both the static body shape and dynamic walking patterns for person Re-ID. Specifically, we digitize multiple 2D AoA images of a person into a 3D human body representation, which is independent of the angle of view. To extract the static body shape, we innovatively use the graph neural network (GNN) to learn the person's static features from thousands of unordered and discrete 3D point clouds of the 3D human body under various poses or activities. Moreover, to better extract dynamic walking patterns, we leverage the gated recurrent units (GRUs) to process joint accelerations from time-series poses of the 3D human body. These features are essentially the static and dynamic biometrics of the human, which are relatively stable over time, enabling a more robust person Re-ID system and more explainable. At last, we use a Siamese network architecture to train the Re-ID network, which can identify people by comparing the similarity between people.

We experimentally evaluate the 3D-ID system with 28 people in various indoor environments including the laboratory, classroom, and home. We perform person Re-ID across different environments, as well as under various attacks, where an adversary tries to evade or deceive the system. We also compare our system to the vision-based and prior WiFi-based person Re-ID systems. Experimental results show that our Re-ID system is highly accurate across time and space. The main contributions of our work are summarized as follows:

- We propose a WiFi vision-based approach for person Re-ID in 3D space. Our approach leverages multiple antennas on next-generation WiFi devices and the 2D AoA estimation of the signal reflections to visualize the person.

- We propose to extract intrinsic and persistent static body shape and dynamic walking patterns of a person to perform person Re-ID. Thus, our system is more robust and has better explainability than prior WiFi-based systems.

- Extensive experiments in various indoor environments demonstrate that our system can achieve rank-1, rank-2, and rank-3 accuracies of 85.3%, 91.7%, and 96.3%, respectively. Moreover, our system is resistant to various attacks.

## 2 Related Work

We categorize existing work in surveillance and person Re-ID/identification into three categories: computer vision-based approaches, RF sensing-based approaches, and others.

**Computer Vision-based.** Many computer vision-based intelligent surveillance and person Re-ID/identification systems have been proposed due to the advancement of deep learning algorithms and the increasing availability of vision datasets. For example, some systems can leverage images with a deep learning network [1, 6, 22, 32, 40, 77] to identify different people. More specifically, given a pair of im-ages as input, these systems output a similarity value indicating whether the two input images depict the same person. Some research efforts [2, 35, 42, 68] further developed spatial-temporal network architectures for video-based person Re-ID, which utilizes a convolutional neural network (CNN) with recurrent neural network (RNN) layers to capture both appearance and motion information of the person. Moreover, recent vision-based works [5, 79] explore the prior knowledge of the 3D body structure which can be recovered from 2D images or videos [27] and improve the robustness of person identification by using the information of a person's 3D shape. However, computer vision-based systems cannot work in NLoS or poor lighting scenarios. The changes in a person's poses and clothes can also mislead the identification [15]. Furthermore, vision-based techniques frequently incur a non-negligible cost [51].

**RF Sensing-based.** In recent years, researchers have leveraged radio frequency (RF) sensing-based techniques for various sensing tasks, such as large-scale activity sensing [61, 64], small-scale motion sensing [37, 60], indoor localization [36, 57, 78], object sensing [50, 62], and human pose estimation [51, 52]. Furthermore, researchers demonstrate the possibility of utilizing RF signals to conduct person Re-ID and identification. For instance, Fan et al. [15] proposed RF-ReID, a person Re-ID system using FMCW radio signals. The FMCW radio can traverse clothes and reflect off the human body. Therefore, it can extract persistent features (e.g., body shape) for human identification. This system, however, relies on specialized hardware that transmits FMCW signals over a wide bandwidth that is dozens of times greater than that of WiFi. It also needs a carefully designed and synchronized T-shape antenna array. Meanwhile, many systems utilize WiFi devices to enable potential mass adoption as WiFi devices are pervasive and can be found in a home environment [59]. There has been work to use WiFi signals for eavesdropping keystrokes [16] and voice liveness detection [44]. WiWho [71] is a framework for identifying a person using the gait information detected via WiFi signals in the time domain. WiFi-ID [72] both identify a person's walking steps and gait patterns by extracting statistical information from WiFi signals in the frequency domain. Instead of using handcraft features, more and more WiFi-based systems leverage deep neural networks to boost person identification. Shi et al. [56] utilized CNN to extract unique human behavioral characteristics inherited from their daily activities sensed by WiFi signals for identification. Some systems develop user identification using CNN and WiFi signals that can work under various scales of environmental dynamics [55, 73] or multi-user scenarios [29]. However, instead of exploring explainable domain knowledge of humans (e.g., body shape and walking dynamics), these WiFi-based systems use a black-box-based approach to indirectly input the WiFi signals to deep learning networks. We note that Huang et al. [23] did primiparity work to image a simple object using specialized WiFi devices. Particularly, they require the use of the customized device of USRP. And their image resolution is too low to visualize the human body or activities as no spatial or frequency diversity was exploited. We also note that our system is different from the traditional radar systems [41]. The radar systems utilize expensive specialized hardware to generate high-energy signal beams and are not designed for indoor environments.

**Other Approaches.** Other person identification approaches could leverage physiological biometrics like fingerprint [8], face [17, 31], voice [24, 43, 74, 75], signature [48, 49], ear canal [67], and teeth [66] to identify the person. However, dedicated biometric hardware and user involvement are required for these approaches, which incur extra costs. Moreover, some wearable sensor-based approaches [46, 47] require a user to wear or carry various sensors and collect gait information for person identification. For example, Ren et al. [46] leveraged the accelerometer embedded in mobile devices to record distinctive gait patterns for user identification. Nevertheless, wearable sensor-based systems require the user to carry or wear one or more physical sensors, which can be intrusive, inconvenient, and cumbersome due to the explicit involvement of users.

## 3 Preliminaries

### 3.1 WiFi Sensing Basics

The prevalence of WiFi networks in public and at home provides us the opportunity to utilize the pervasive WiFi signals to sense humans and interpret activities. Similar to visible lights, WiFi signals travel through space, reflect from the human body and physical objects, and undergo wave phenomena such as diffraction. The WiFi signals thus capture a considerable amount of information about the environment including humans and their activities. Unlike cameras that use CCD/CMOS sensors to sense the intensity of visible lights at the megapixel resolution, older versions of WiFi devices equipped with one omnidirectional antenna can only capture one received signal strength (RSS) per packet, which characterizes the overall energy of the signals reflected from the entire environment. This basically means it can only produce one "pixel" value when compared to that of an optical image. Still, based on the change of a single "pixel", certain applications can be built, such as intrusion detection in restricted areas [70].

The current WiFi standard employs OFDM technology to provide fast and reliable communication. It partitions each WiFi channel into multiple OFDM subcarriers and transmits data on each of the subcarriers. It thus provides channel state information (CSI), which contains amplitude and phase measurements separately for each OFDM subcarrier. For example, on a standard 20MHz channel, WiFi radios measure amplitude and phase for each of the 56 OFDM subcarriers. This means that current commodity WiFi devices could provide up to 56 "pixel" values per packet by leveraging CSI.

The CSI exported from the commodity WiFi devices only provides information on how each OFDM subcarrier was interrupted by humans and their activities. It however offers no explicit spatial information regarding human body shape or poses. Existing WiFi sensing systems mainly employ a black-box approach, in which deep learning techniques are utilized to model the relationship between the CSI changes and human activities. They assume similar activities or persons will result in similar signal change patterns. This approach may work for a limited number of pre-defined activities from controlled users, for example, to distinguish family members with several pre-defined activities for controlling smart home appliances. However, it lacks reliability and is less applicable to security-sensitive applications as one signal change pattern may correlate with an infinite number of free-form activities or uncontrolled users. Moreover, the black-box-based approach is also subject to malicious attacks, where an adversary could manipulate WiFi signals to bypass the person Re-ID systems [28, 73]. This type of attack has already been demonstrated in recent WiFi-based person identification systems [38]. Therefore, existing WiFi-based person identification methods cannot be directly leveraged to build a secure and robust person Re-ID system.

## 3.2 2D AoA-based WiFi Sensing

As the next generation of WiFi supports a large number of antennas at each WiFi device (e.g., WiFi 7 supports up to 16 antennas), we leverage multiple spatially separated antennas to estimate the direction of the signal reflections to visualize a person in the environment. Given the intensity of the signal reflections in each direction, we could derive a visualization of the person in the environment similar to that of an optical grayscale image. Thus, the fairly large number of antennas on next-generation WiFi devices become an enabler for generating 2D AoA-based visualizations for person Re-ID. It enables the WiFi devices to "see" the visual world as humans do.

Let's assume the multiple antennas on a WiFi receiver form an L-shaped distribute antenna layout (array), as shown in Figure 1. We note that the L-shaped antenna array is among the best to estimate 2D AoA in terms of azimuth and elevation angles [19]. Specifically, the azimuth ($\varphi$) is an angular measurement of the signal on the horizon, whereas the elevation ($\theta$) is the angular measurement of the same signal in the vertical direction. Then, the direction of an incident signal to the receiver in physical space (e.g., incident signal 1 in Figure 1) can be uniquely determined by azimuth and elevation (i.e., 2D AoA).

We assume that there are $S$ incident signals and $K$ antennas in Figure 1. For simplicity, we omit the line-of-sight (LoS) signal as well as other signal reflections from the environment. That is, we only illustrate the signals reflected from the human body (i.e., incident signals) in Figure 1. These signals are arriving at the antenna array in the $X - Y - Z$ coordinates. The phase shift for the $s^{\text{th}}$ incident signal on the $k^{\text{th}}$ antenna can be denoted as:

$$\Phi_k(\varphi_s, \theta_s) = e^{-j2\pi f \cdot \mathbf{d}_s(\varphi_s, \theta_s) \cdot \mathbf{l}_k / c} \quad (1)$$

where $c$ is the speed of light, $\mathbf{l}_k = [x_k, y_k, z_k]^T$ is the position vector of antenna $k$ that depends on the geometry of the antenna array (in our system $y_k = 0$) and $\mathbf{d}_s(\varphi_s, \theta_s)$ is unit vector pointing towards incident signal $s$. Assuming that the incident signal arrives from azimuth $\varphi$ and elevation $\theta$, we have $\mathbf{d}(\varphi, \theta) = [\cos(\varphi)\sin(\theta), \sin(\varphi)\sin(\theta), \cos(\theta)]$. Therefore, the phase shifts for the $s^{\text{th}}$ incident signal across all $K$ antennas in a given antenna array can be written as:

$$\mathbf{a}(\varphi, \theta) = [\Phi_1(\varphi_s, \theta_s) \, \Phi_2(\varphi_s, \theta_s) \, ... \, \Phi_M(\varphi_s, \theta_s)]^T. \quad (2)$$

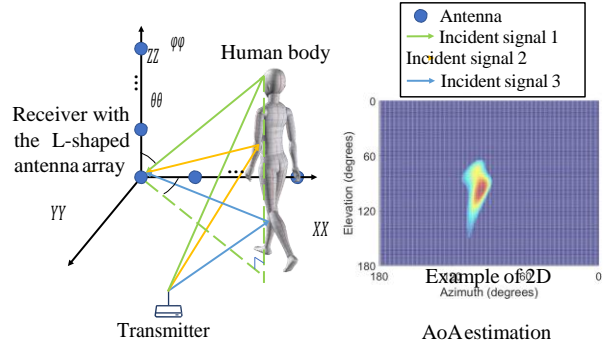

Figure 1: 2D AoA-based WiFi sensing.

Here, $\mathbf{a}(\varphi, \theta)$ is also called the steering vector for 2D AoA. Thus, the steering matrix for all $S$ incident signals is

$$\mathbf{A}(\varphi, \theta) = [\mathbf{a}(\varphi_1, \theta_1), \ldots, \mathbf{a}(\varphi_S, \theta_S)]. \quad (3)$$

Given the steering matrix $\mathbf{A}(\varphi, \theta)$, the azimuth angle and elevation angle of each incident signal can be derived by using the MUSIC algorithm [54] and shown as the peaks in the 2D AoA spectrum. We refer to the estimated 2D AoA spectrum as an "image" or visualization in our work.

The example of the 2D AoA image is shown in the right part of Figure 1, in which the horizontal axis represents the azimuth, the vertical axis represents the elevation and the color represents the signal power. We can observe the shape of a walking person in the spectrum. However, compared with images generated by a camera that have thousands of pixels or even megapixels, the spatial resolution provided by multiple antennas is still very limited. Therefore, we cannot distinguish different body parts of a person such as the head, torso, and limbs, nor can we observe the pose of the person. Thus, such a resolution is insufficient for person Re-ID and requires further improvements, which will be illustrated in Section 5.2.1.

# 4 System and Threat Model

## 4.1 System Model and Applications

Our system requires multiple antennas to sense and identify people. However, there is no specific requirement on the antenna layouts. Although the L-shaped distributed antenna layout achieves the best performance, any antenna layout could work including circular layout, rectangular layout, etc. Our system could be a software update using existing WiFi networks that can export CSI such as Intel and Atheros NICs. More NICs vendors are open to exporting CSI as a new standard in IEEE 802.11, known as IEEE 802.11bf [13], is being finalized to enhance sensing capabilities through 802.11-compliant waveforms.

The Re-ID system aims to identify people across different times and locations. Our system could support various security applications and may also augment traditional camera-based systems. Such a person Re-ID system can enable security monitoring. Assuming a shop was stolen and the thief escaped. When the thief shows up elsewhere, the system will identify him/her and raise alerts. Also, person Re-ID can be applied in public areas for person tracking. If one person (e.g., a child) is lost in a large shopping mall, we can use the WiFi devices around the mall to track and find the person even if the person is blocked by some obstacles. Moreover, the Re-ID system could be used for access control. The system can identify people outside the door. If a person is not a legitimate user and approaches the property, the system will raise alerts or prevent unauthorized access.

## 4.2 Threat Model

In the threat model, an adversary aims to evade or deceive our Re-ID system. That is to say, the adversary is to avoid being captured by the system or to cause the system to retrieve a person of the wrong identity. The attacks can be divided into two categories: naive attacks and advanced attacks. For naive attacks, the adversary launches attacks on our system in the same way as attacking vision-based systems. They include (a) the adversary can change clothes (e.g., color and style) which leads to the change of appearance; (b) the adversary can hide behind an obstacle that could result in the NLoS scenarios. For advanced attacks, the adversary has some knowledge about our Re-ID system and can launch sophisticated attacks including (c) the adversary can deliberately change the walking behaviors; (d) the adversary can generate wireless interferences to disrupt the WiFi channel.

# 5 System Design

## 5.1 System Overview

The key idea of our WiFi vision-based approach for person Re-ID is to leverage the advances of WiFi and deep learning to help WiFi devices "see", identify, and recognize people. As illustrated in Figure 2, the system takes as input time-series CSI measurements at multiple antennas of the WiFi receivers when a person is walking around in the environment. The system can reuse existing WiFi devices and take advantage of CSI measurements from existing traffic, or if insufficient network traffic is available, the system might also generate periodic traffic for measurement purposes. The WiFi signals reflected from the human body (e.g., head, torso, arms, and legs) will travel through different paths and arrive at the receiver in various directions (i.e., azimuth and elevation angles). The CSI measurements of received signals then go through the 2D AoA estimation component to calculate the azimuth and elevation angles of the signal reflections. Next, our system performs human enhancement to segment signal reflections of the human body from irrelevant reflections of the surrounding environments. It then performs multi-frame aggregation to fully capture signal reflections of the human body.

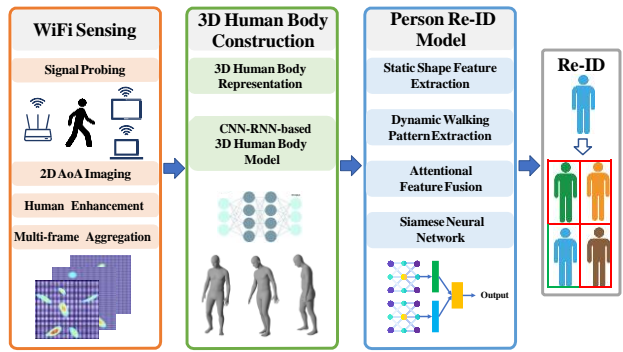

Figure 2: System overview of 3D-ID.

After the WiFi devices "see" a human, we need to extract static body shape and dynamic walking patterns independent of the angle of view. Thus, we leverage deep learning models to digitize the 2D human images into a 3D human body representation. Specifically, our system constructs a 3D human body based on the Skinned Multi-Person Linear model (SMPL) [39]. Once we obtain the time-series 3D human body, our system extracts persistent and long-term features including static body shape and dynamic walking patterns for person Re-ID. We leverage a GNN-based method to extract the static body shape features from the 3D human body. We choose a GNN-based method because the surface of the 3D body consists of unordered and discrete 3D point clouds that cannot be processed by traditional CNN but can be considered as a graph and handled by GNN. Meanwhile, we calculate the acceleration of each joint of the 3D body and take it as input to multi-layer GRUs to extract dynamic walking patterns. After that, our system fuses both static body shapes and dynamic walking patterns via an attentional feature fusion method. At last, we use a Siamese network architecture to train the neural network to perform person Re-ID.

Our proposed system could leverage the prevalence of WiFi signals in public and at homes and reuse pervasive WiFi devices for person Re-ID. It thus presents tremendous cost-saving when compared with dedicated camera-based systems. As the WiFi signals can traverse occlusions and clothes and can illuminate the human body, the proposed 3D-ID system can work under NLoS or poor lighting conditions and is less affected by an individual's appearance. Our system leverage domain knowledge of human (i.e., both static body shape and dynamic walking features), which makes our system more robust and explainable than prior black-box WiFi systems.

## 5.2 WiFi-based Human Body Imaging

In this subsection, we describe the details of the 2D AoA-based human imaging, which is the foundation of our proposed WiFi vision-based person Re-ID.

### 5.2.1 2D AoA Imaging

To achieve 2D AoA-based human body imaging, we utilize four-dimensional information: time diversity of multiple WiFi packets, spatial diversity of both receiving and transmitting antennas, and the frequency diversity of OFDM subcarriers. We first leverage a time sequence of WiFi packets (i.e., time diversity) to obtain a signal matrix. By using more WiFi packets, the estimation variance for the covariance matrix in the MUSIC algorithm will decrease [33], which will result in sharper peaks in the derived AoA spectrum. By doing this, it is easier to distinguish various signal reflections from different subjects or body parts, thus improving the quality of the 2D AoA-based image. In this work, we empirically leverage 100 WiFi packets for the time diversity of WiFi packets since using a larger number will lead to a longer processing delay.

The resolution of an optical image depends on the number of pixels produced by the CCD/CMOS sensor. Similarly, the resolution of the 2D AoA-based image depends on the number of antennas the WiFi receiver used to receive and discern the signal reflections. However, we cannot arbitrarily increase the number of antennas on the WiFi receiver as this number is normally limited. For example, the next generation WiFi 7 only supports up to 16 antennas. Instead of increasing the antennas on the WiFi receiver, we can leverage the multiple antennas on the WiFi transmitter to improve the resolution of the 2D AoA estimation. This is because the spatial diversity in transmitting antennas can also create a phase difference at each of the receiving antennas. Thus, we can incorporate the spatial diversity in both transmitting and receiving antennas to further improve the resolution of the 2D AoA spectrum. We assume that the WiFi signals are emitted from a linear antenna array at the transmitter and will be received with a phase shift $\Gamma(\omega)$. Note that $\Gamma(\omega)$ is the function of angle of departure (AoD) $\omega_s$. Thus, the phase difference across transmitting antennas can be written as:

$$\Gamma(\omega_s) = e^{-j2\pi f \cdot d \sin(\omega_s)/c}, \quad (4)$$

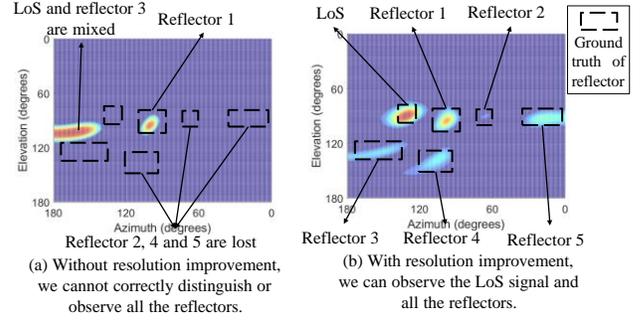

Figure 3: The effect of increasing the resolution. The environment includes the LoS and five main reflectors.

where $s$ is the index of the signal path and $d$ is the distance between two adjacent transmitting antennas.

Moreover, the current IEEE 802.11 standard using OFDM technology transmits data over multiple subcarriers. Therefore, we can leverage the frequency diversity of OFDM subcarriers to further improve the resolution as the frequency diversity creates a phase difference across different frequencies as well. Specifically, for the $s^{th}$ path, we denote the time of flight (ToF) as $\tau_s$. Such a ToF will introduce a phase shift across two consecutive OFDM subcarriers. Assuming the frequency difference between two OFDM subcarriers is $f_\delta$, the phase difference across OFDM subcarriers can be denoted as:

$$\Omega(\tau_s) = e^{-j2\pi f_\delta \cdot \tau_s/c}. \quad (5)$$

Now, we can jointly estimate the 2D AoA (azimuth and elevation) with ToF and AoD by constructing a large virtual antenna array from all subcarriers of all receiving antennas for signal streams transmitted from all transmitting antennas. This information can be obtained from WiFi NICs with MIMO-OFDM techniques. Let $N_{Rx}$, $N_{Tx}$, and $N_{Su}$ be the numbers of receiving antennas, transmitting antennas, and subcarriers, respectively. The CSI measurement in the format of $N_{Rx} \times N_{Tx} \times N_{Su}$ represents the overall phase shift and attenuation introduced by the channel measured at each virtual antenna. Therefore, the virtual antenna array can be constructed using CSI from all the subcarriers at all of the physical antennas.

Compared to the previous steering vector in Section 3.2, we have a new steering vector $\mathbf{a}(\phi, \theta, \tau, \omega)$ which is formed by phase difference introduced at each virtual antenna and it can be denoted as:

$$\mathbf{a}(\phi, \theta, \tau, \omega) = [\mathbf{a'}(\phi, \theta, \tau), \Gamma_\omega \mathbf{a'}(\phi, \theta, \tau), \dots, \Gamma_\omega^{N_{Tx}-1} \mathbf{a'}(\phi, \theta, \tau)]^T, \quad (6)$$

where $\Gamma_\omega$ is the abbreviations of $\Gamma(\omega)$ and $\mathbf{a'}(\phi, \theta, \tau)$ can be written as:

$$\mathbf{a'}(\phi, \theta, \tau) = [1, \dots, \Omega_\tau^{N_{Su}-1}, \Phi_{(\phi,\theta)}, \dots, \Omega_\tau^{N_{Su}-1} \Phi_{(\phi,\theta)}, \dots, \Phi_{(\phi,\theta)}^{N_{Rx}-1}, \dots, \Omega_\tau^{N_{Su}-1} \Phi_{(\phi,\theta)}^{N_{Rx}-1}]^T, \quad (7)$$

where $\Omega_\tau$ and $\Phi_{(\phi,\theta)}$ are the abbreviations of $\Omega(\tau)$ and $\Phi(\phi, \theta)$, respectively.

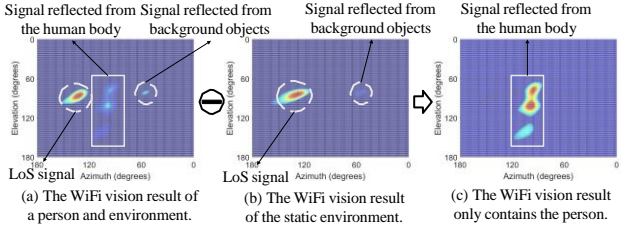

Figure 4: Results of human image enhancement.

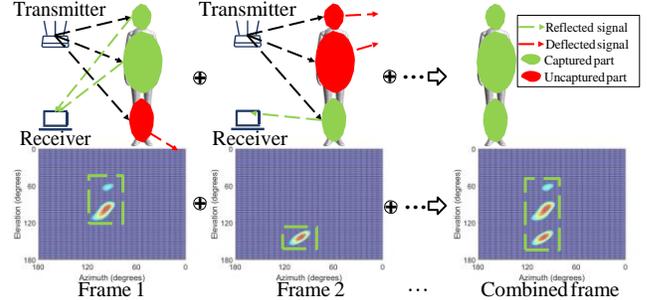

Figure 5: An intuitive example of multi-frame aggregation to capture the whole human body.

Therefore, we can construct the received signal using the above new steering vector. Parameters of 2D AoA, ToF and AoD that maximize the spatial spectrum function [54] can be given by:

$$P(\phi, \theta, \tau, \omega)_{Improve} = \frac{1}{a^H(\phi, \theta, \tau, \omega) E_N E_N^H a(\phi, \theta, \tau, \omega)} \quad (8)$$

where $E_N$ represents the noise subspace. At last, we derive the 2D AoA spectrum as an image by superposing the dimensions of ToF and AoD.

It is worth noticing we need to perform CSI de-noising to clean the CSI phase noises before 2D AoA estimation. This is because the hardware imperfections of WiFi devices could lead to CSI phase distortions. In particular, the receiver suffers from a random phase shift caused by the sampling time offset (STO) and packet detection delay (PDD) across packets. STO and PDD have a constant effect across all transmitting or receiving antennas since all the radio chains of WiFi NICs are time-synchronized. To sanitize such random phase offsets with each WiFi NIC, we apply a linear fit method proposed in [30] to unwrapped phase measurements for the two WiFi NICs respectively.

We show an example in Figure 3 to illustrate the effect of resolution improvement. For this figure, our setup has the receiver to receive the LoS signal and there are five main reflectors involved. The ground truths of their sizes and locations are also shown in the figure. We first performed 2D AoA estimation with only 1 WiFi packet, 9 receiving antennas, 1 transmitting antenna, and 1 subcarrier. The results are shown in Figure 3(a). We can observe that the majority of signal reflections are either indistinguishable from each other or completely indiscernible. For example, the LoS signal and signal reflections caused by reflector 3 are mixed together. Signal reflections caused by reflectors 2, 4, and 5 are not distinguishable from the surrounding environments. We then performed resolution improvement by using 100 WiFi packets, 9 receiving antennas, 3 transmitting antennas, and 30 subcarriers. The result is shown in Figure 3(b). We can clearly observe the signal reflections from different reflectors in the 2D AoA spectrum, which matches the ground truth.

### 5.2.2 Human Image Enhancement

As shown in Figure 4(a), similar to an image captured by a camera, the generated 2D AoA from the previous step provides spatial information for all reflections which mainly contains the signals reflected from the human and the environment. However, in person Re-ID, we are only interested in the human. The static environment could be considered background noises. Thus, we propose to perform human image enhancement to highlight the human while mitigating the impact on the surrounding environment. Therefore, only the human-dependent reflections will be retained.

Inspired by the spectral subtraction method in digital processing [4], which involves subtracting the estimated noise spectrum from the image, we can enhance the human by subtracting the static component/spectrum in a sequence of 2D AoA images. In this way, the enhanced image will mostly capture the signals bounced off the human body, and the signals reflected from the surrounding environments will be removed. For example, Figure 4(a) shows the 2D AoA image when the person is walking in the same environment, whereas Figure 4(b) shows the signals of the static component of the image including the signals reflected from the background objects (e.g., desk) and the LoS signal. The LoS signal traveled through a shorter distance and therefore is much stronger than the signals reflected from the human body. This makes it very difficult to observe the person of interest. After we perform the subject enhancement, we can clearly observe the enhanced signal reflections of the person in Figure 4(c), in which different parts of the body clearly stand out from the surrounding environment.

We note that the signals reflected from the human body may bounce off the static environment again, and such secondary reflections may be also captured by the receiver. For example, when a person walks in the environment, the signal reflected from the person may bounce to the wall and later be captured by the receiver. Such signal reflections have very little impact on 2D AoA imaging and can be removed by applying a threshold since the signal reflected by the human body is much stronger as it experiences less attenuation.

### 5.2.3 Multi-frame Aggregation

With respect to WiFi signals, the human body is considered specular, which means that the human body acts as a reflector

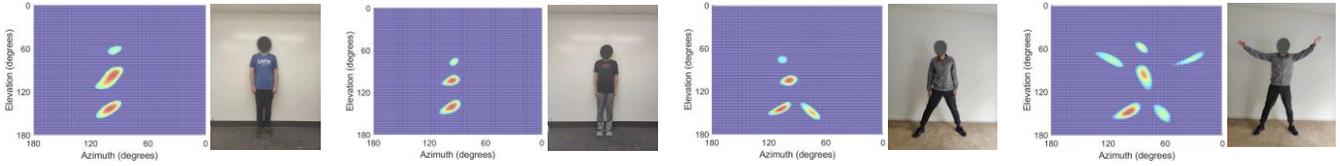

(a) A person with a larger body shape and the corresponding WiFi vision result.

(b) A person with a smaller body shape and the corresponding WiFi vision result.

(c) A person is splitting legs and the corresponding WiFi vision result.

(d) A person is lifting both arms and the corresponding WiFi vision result.

Figure 6: Results of WiFi vision for various people and activities.

(i.e., a mirror) rather than a scatterer. This is due to the fact that the wavelength of the WiFi signal is significantly longer than the roughness of the human body's surface [76]. Conversely, the human body scatters visible light as its wavelength is much shorter. The WiFi signal may be reflected towards or away from the receiver depending on the direction of the surface for different body parts. Thus, unlike images generated by camera systems, in which a single image frame can capture all unobscured parts of the human body, a single 2D AoA image frame only captures a subset of the human body while missing the parts that deflect the WiFi signals.

As shown in Frame 1 of Figure 5, the WiFi signals reflected from the head and torso are captured by the receiver while the legs deflect the signal away from the receiver. Thus, the first frame of the image only captures the upper and middle parts of the human body and does not capture the low part. On the contrary, Frame 2 captures the lower part of the human body but fails to capture other body parts, as shown in Figure 5. To resolve this issue, we can combine multiple frames of images to capture a more complete picture of the human body. Specifically, the design of our deep learning network in later sections takes into account the specularity of the human body and aggregates information from multiple (e.g., 15) image frames.

After multi-frame aggregation, our system can obtain recognizable, enhanced, and complete 2D AoA images for a person, as shown in Figure 6. In particular, Figure 6(a) and (b) show the 2D AoA images of two people with different body shapes. We can observe that a larger body shape creates a larger area of signal reflections in the image. Figure 6(c) and (d) show that the person is performing different activities and the corresponding 2D AoA image can display the poses of that person. To summarize, the 2D AoA image is capable of revealing information about both body shape and pose, thus providing the foundation for WiFi vision-based person Re-ID.

### 5.3 3D Human Body Construction

In this subsection, we describe the process of digitizing the human body into a 3D representation based on multiple 2D AoA images and a deep learning network. Through this process, we can obtain fine-grained domain knowledge of body shape and walking patterns for person Re-ID.

To generate a realistic 3D human body, we use the Skinned Multi-Person Linear model (SMPL) [39] as our 3D human body model. SMPL is a widely used parametric model that estimates the 3D human body by encoding the human body into pose and shape information. SMPL outputs a triangulated mesh with 6890 vertices which can represent people's height, weight, and body proportions. Also, the SMPL model can represent different human poses with coordinates of 24 joints.

The deep learning model for 3D human body construction is shown in Figure 7. It is worth noting that we utilize two receivers in orthogonal directions to generate multi-view 2D AoA images and construct the human body in 3D space. Thus, we take the tensor with the dimensions of $15 \times 2 \times 180 \times 180$ as input of the network, where 15 is the number of frames, 2 is the number of receivers and 180 means the range of azimuth or elevation angles (range of the angle is $[1, 180]$ degrees with a step of 1 degree).

In the model architecture, we first adopt CNN to extract spatial features from the 2D AoA images. It is because these images contain the general shape and pose information of the human body, CNN can help the system map such information to different vertices of the body. In particular, we use ResNet-18 which is a widely used CNN architecture [20]. After the CNN layer, we utilize a max-pooling layer to extract the most relevant features and remove redundant information. We then use a two-layer GRU as the recurrent layer which can model a sequence of temporal dynamics. Because each frame has a distinct effect on the result over time, we use a self-attention technique [27] to dynamically learn the relative contributions of each frame and emphasize the most relevant frames' contributions in the final representation. At last, we map the output of the self-attention layer to the shape and pose parameters in the SMPL model to generate the 3D human body. We consider the training of the model as a regression problem that minimizes the error of both shape and pose parameters of SMPL. Let $\alpha$ and $\beta$ be the ground truth of pose and shape parameters, respectively. $\hat{\alpha}$ and $\hat{\beta}$ represent the predicted pose and shape parameters, respectively. The loss function is a weighted sum of pose loss as well as shape loss which can be written as:

$$L_{Human} = \sum_{t=1}^{T} a \cdot ||\alpha_t - \hat{\alpha}_t||_{L_1} + b \cdot ||\beta_t - \hat{\beta}_t||_{L_1}$$

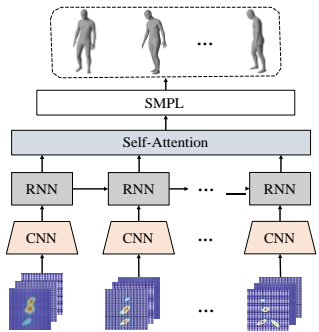

Figure 7: The 3D human body model.

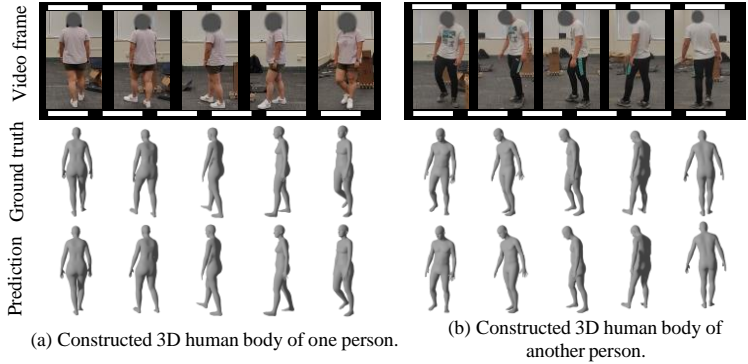

(a) Constructed 3D human body of one person.

(b) Constructed 3D human body of another person.

Figure 8: Examples of the constructed 3D human body for different people.

where T is the number of frames, $L_1$ means $L_1$ norm, and $a, b$ are the weights to balance the losses.

Examples of constructed 3D human bodies for two people who are walking are shown in Figure 8. The top row shows the time series video frames recorded by an RGB camera for visual reference. The middle row shows the ground truths of 3D human bodies which are generated by the state-of-the-art computer vision-based approach [27]. The last row illustrates the 3D human bodies constructed by our model. We can observe that the 3D human bodies constructed by our system are almost identical to the ground truth. Moreover, we observe both the shape and walking pose of the constructed 3D human body are different for the two people in Figure 8, which provide the domain knowledge of human for person Re-ID.

## 5.4 Person Re-ID Model

We build a novel network to extract both static shape and dynamic walking patterns from constructed 3D human bodies for person Re-ID. As shown in Figure 9, we extract static features from the body shape by using a GNN-based method, which considers the 3D body as a graph and produces a vector representing the person's static shape biometrics. The dynamic walking pattern features contain the gait pattern, arm and torso gestures, head and body movements. Thus, we compute the acceleration of each of the 24 joints based on each 3D human body. Then, we make acceleration information flow between time steps by using a recurrent layer and all time steps are combined using temporal pooling. Next, these static shape and dynamic walking pattern features are fused together using an attention-based mechanism. Lastly, the two-stream sub-networks for two 3D body sequences from two different people are constructed following the Siamese network architecture, in which the parameters of sub-networks are shared. Given a pair of 3D body sequences from the same person, the Siamese architecture is trained to produce feature vectors that are close in feature space, while given a pair of 3D body sequences from different persons, the network is trained to produce feature vectors that are separated.

### 5.4.1 Static Shape Feature Extraction

Extracting static shape biometrics of the person from the 3D point clouds of 3D bodies is challenging. The reason is that the 3D body consists of thousands of 3D point clouds, which are discrete and unordered when compared to the image format [79]. Therefore, we cannot utilize the typical 2D convolutional operation to extract the static shape biometrics. To overcome this challenge, we leverage the idea of GNN [53] to construct a graph according to the distance between points. Here, we adopt a dynamic graph convolution network (GCN) to learn from the constructed graph.

In our work, the k-nearest neighbor graph $\mathcal{G} = (\mathcal{V}, \mathcal{E})$ is used to model the relationship between neighbor points, where $\mathcal{V}$ means the vertex set, and $\mathcal{E}$ means the edge set. The k-nearest neighbors are chosen based on the points value. Then, we use dynamic graph convolution [65] to learn representation from the topology structure of the graph. Let $\mathbf{x_p}$ be the point feature (e.g., 3D coordinates) for point $p$ and the output of the dynamic graph convolution can be denoted as $\mathbf{x'_p} = \sum_{q:(p,q)\in \mathcal{E}}(\varepsilon_p \mathbf{x_p} + \varepsilon_q \mathbf{x_q})$, where $\mathbf{x_q}$ means the feature of neighbor points, and there is one edge from $p$ to $q$. $\varepsilon$ is the learnable parameter. After the dynamic graph convolution layer, we also use the batch normalization layer and rectified linear unit to obtain the static shape features.

### 5.4.2 Dynamic Walking Pattern Extraction

In order to extract dynamic walking pattern features including gait, arm, body, and head movement patterns, we calculate the acceleration for each joint [39] (24 joints in total in SMPL) based on the joint positions in the 3D human body. We then take accelerations as input feed into a two-layer GRU since our system needs to recognize a person using a sequence of bodies. Recurrent connections can help this process by allowing information to be passed between time steps.

Although GRUs are capable of capturing temporal information, the output may be biased towards later time steps. To resolve this issue, we leverage a temporal pooling layer [42], which enables the aggregating of information over all time steps, thus preventing bias towards later time steps. The temporal pooling layer can collect the long-term information of sequence. Moreover, such a layer can capture long-term information in the sequence and aims to represent information at all temporal scales within the input.

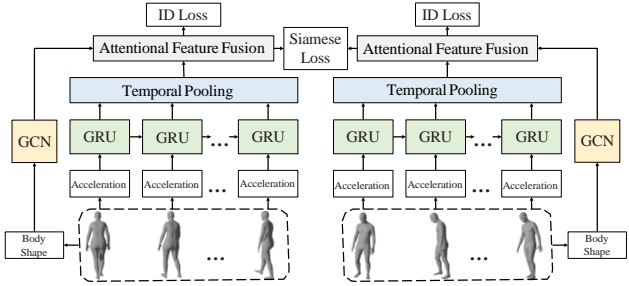

Figure 9: The person Re-ID model.

#### 5.4.3 Attentional Feature Fusion

The static shape and dynamic walking pattern features need to be fused for the person identification. However, traditional feature fusion is usually implemented with simple operations such as summation or concatenation. But those approaches are not optimal for our system. Thus, we leverage the attentional feature fusion to combine the two feature maps. Let **S** and **D** be the static shape and dynamic pose feature maps, respectively. The fusion of two feature map can be written as $\mathbf{F} = \mathbf{M}(\mathbf{S} \uplus \mathbf{D}) \otimes \mathbf{S} + (1 - \mathbf{M}(\mathbf{S} \uplus \mathbf{D})) \otimes \mathbf{D}$, where **M** is a multi-scale channel attention module [10] and $\uplus$ means the initial feature integration. The key point of fusion is that the network can use the fusion weights to perform a soft selection or weighted averaging between two types of features.

#### 5.4.4 Siamese Neural Network

The proposed person Re-ID network can be trained using the Siamese network architecture [7], which includes two sub-networks with the same weights. The purpose of the Siamese network is to find a similarity or a relationship between two comparable objects. Once the Siamese network is trained, it can compare the body shapes and walking patterns of two people and thus can determine if they are the same person. When a pair of inputs are given to the network, the sub-networks map them to a pair of feature vectors, which are then compared using Euclidean distance. For our person Re-ID system, we want to map body sequences from the same person to feature vectors that are close and map body sequences from different people to feature vectors that are widely separated. By computing the distance between the identity feature vectors with feature vectors of other identities, the lowest distance indicates the most similar identity.

Given the fused feature vectors $(F_i, F_j)$ for person $i$ and person $j$, we utilize the Euclidean distance Hinge loss to train our model:

$$L_{Siamese}(F_i, F_j) = \begin{cases} ||F_i - F_j||^2, & i = j \\ \max(0, \sigma - ||F_i - F_j||^2), & i \neq j, \end{cases} \quad (10)$$

where $\sigma$ is the margin to separate features. We also calculate the identity loss $L_{ID}(F_i)$ and $L_{ID}(F_j)$ using the cross-entropy loss. The final loss function is written as $L_{ReID} = L_{Siamese} + L_{ID}(F_i) + L_{ID}(F_j)$.

## 6 Performance Evaluation

### 6.1 Experimental Setup

**Devices.** We conduct experiments using one WiFi transmitter and two receivers. Specifically, the WiFi transmitter is equipped with a linear antenna array of three antennas. Each receiver is equipped with an L-shaped antenna array of nine antennas as shown in Figure 10. Note that we stitch NICs with shared antennas using splitters to simulate the antenna configuration of the new generation of WiFi devices. Linux 802.11 CSI tools [18] are used to extract CSI measurements from 30 subcarriers with a bandwidth of 40 MHz. The default packet transmission rate is set at 1000 packets per second. We utilize a camera to record the ground truth for both the 3D body and Re-ID for the person. We use network time protocol (NTP) to ensure synchronization for all devices.

**Environments.** We evaluate our system in four different environments including two laboratories, a classroom, and a living room. As shown in Figure 11, the size of both laboratories is $4.5m \times 4.5m$. The two labs have different furniture setups. The sizes of the classroom and living room are $8.5m \times 5.5m$ and $6m \times 6m$, respectively. The deployments of all devices in each environment are also described in Figure 11. If not specified, the default distance between the transmitter and receivers is $2m$. We note that the person can walk freely in WiFi environments. People occasionally change clothes and the lighting conditions may vary in experiments.

**Model Settings.** For the 3D human body construction network, we use the ResNet-18 framework as the feature extractor. We utilize 2 layers of GRU as the residual model, and the number of the hidden state is set as 2048. The dropout rate is 0.5. We use two fully-connected layers of size 2048 and $tanh(\cdot)$ for the self-attention module. The batch size is set to 16 and the learning rate is 0.0001.

For the person Re-ID network, we also implement a 2-layer GRU. We set the hidden state number as 1024 and the dropout rate as 0.2. The GNN part consists of a graph convolutional network, a rectified linear unit (ReLU), and a batch normalization layer. The network was trained with a learning rate of 0.0001, and a batch size of one. We implement both networks in PyTorch. Our models are trained with the Adam optimizer using the NVIDIA RTX 3090 GPU.

**Dataset.** In this work, 28 participants (20 males, 8 females) of varying heights, weights, and ages were recruited for our

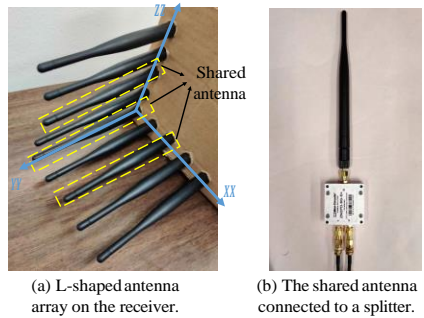

(a) L-shaped antenna array on the receiver.

(b) The shared antenna connected to a splitter.

Figure 10: The constructed L-shaped antenna array on each receiver.

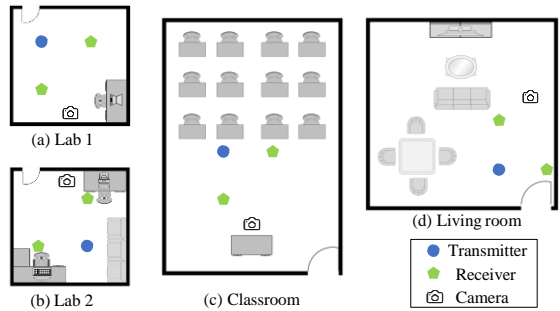

(a) Lab 1

(b) Lab 2

(c) Classroom

(d) Living room

● Transmitter
● Receiver
◉ Camera

Figure 11: Experimental environments and setup.

experiments. Our experiments have obtained IRB approval. We collected synchronized WiFi and vision data of random everyday activities of these participants over two months time period. We have collected over 70 million WiFi packets in total. Moreover, we ask these 28 participants to walk randomly for person Re-ID. Note that each person can walk in each environment at different times and days. For calculating ranking accuracy, the whole dataset is randomly split into two non-overlapping parts: 50% of people (i.e., 14) for training and the remaining 50% of people (i.e., 14) for testing. The experiments are repeated 10 times with different training and test splits and the results are averaged to ensure stable results.

**Baselines.** To demonstrate the effectiveness of 3D-ID, we compare our work with state-of-the-art WiFi-based and vision-based person identification/Re-ID systems. The comparison is conducted by using the same WiFi CSI data for the WiFi-based systems or using the synchronized RGB images/videos for vision-based systems. For RGB images/videos, we place the camera several meters away from the person to simulate a surveillance camera. The camera can capture the whole human body and the images will be used by computer vision-based systems. We note that about 20% people's clothes were changed or NLoS scenarios occurred during the experiments. The first category is image-based systems which focus on appearance features (e.g., clothes colors and hairstyles) in images. In particular, there are two representative systems: Luo et al. [40] proposed a triplet loss to train the Re-ID model and Hermans et al. [22] designed the BNNeck structure for image-based Re-ID. The second category is video-based systems and we reproduce two typical systems: McLaughlin et al. [42] extracted both appearance and walking features from video tracklets, whereas Liu et al. [35] proposed to jointly learn long-range motion context and appearance from videos. The third category is WiFi-based systems that directly utilize CSI to build gait/walking pattern profiles for identification. We reproduce two classical systems: Zeng et al. [71] calculated statistical features (e.g., mean, median, entropy) of CSI, and Zhang et al. [72] used Continuous Wavelet Transformation (CWT) for CSI processing. The last category is WiFi-based systems that derive time-frequency features to recognize people. There are two most recent state-of-the-art systems: Shi et al. [56] extracted both time (e.g., mean, skewness, kurtosis) and frequency (e.g., Short-Time Fourier Transform (STFT)) features, and Kong et al. [29] developed a CNN-RNN feature extractor for CSI data. Note that all of these systems were reproduced with the same number of people and the same data partition in Section 6.1 to ensure a fair comparison.

**Evaluation Metrics.** We leverage the ranking accuracy to evaluate our system, which is a common evaluation metric for person Re-ID. The system is given a WiFi sample of a test person (i.e., the probe) and only one of the candidates (i.e., the gallery) can match the queried WiFi sample of the person. The system then ranks the candidates based on their distances with respect to the test person. The top-k ranking accuracy is defined as the percentage of cases where the correct test person is ranked among the top-k positions of all the candidates in a test. We report the top-1 to top-5 ranking accuracies in this work. We also show the performance with the average Cumulative Matching Characteristics (CMC) curves which measure the probability of top-k matching.

## 6.2 Overall Performance

We first study the overall performance of our system. The person can appear in one environment and walk into another one or appear in the same environment at different times with different walking trajectories, different appearances, or behind obstacles. We compare our system with the existing vision-based person Re-ID systems, as well as WiFi-based person identification systems. As shown in Table 1, our proposed system slightly outperforms image-based systems and is comparable to video-based systems. In particular, our system has rank-1, rank-2, and rank-3 accuracies of 85.3%, 91.7%, and 96.3%, respectively. This shows our system has slightly higher overall ranking accuracies compared to vision-based systems. It is because the reproduced vision-based person Re-ID systems rely on extracting appearance features and walking patterns from LoS images/videos. Thus, the performance suffers from degradation due to the changes in the appearance of people and NLoS scenarios. On the contrary, WiFi signals can penetrate obstacles and clothes. Our system thus can provide robust person Re-ID even when the person wears different clothes or is under NLoS scenarios.

| Modality | System | Rank-1 | Rank-2 | Rank-3 | Rank-4 | Rank-5 |
|---|---|---|---|---|---|---|
| RGB image: appearance features | Luo et al. [40] | 83.0 | 85.9 | 92.7 | 95.1 | 99.1 |
| | Hermans et al. [22] | 82.2 | 84.5 | 91.0 | 94.9 | 99.0 |
| RGB video: appearance and walking features | McLaughlin et al. [42] | 85.8 | 89.3 | 95.4 | 97.9 | **100** |
| | Liu et al. [35] | **86.2** | 90.2 | 95.5 | 98.5 | **100** |
| WiFi: time-frequency features | Kong et al. [29] | 62.4 | 70.3 | 75.8 | 79.7 | 83.4 |
| | Shi et al. [56] | 63.6 | 71.1 | 75.6 | 80.5 | 82.9 |
| WiFi: CSI | Zeng et al. [71] | 53.8 | 58.4 | 64.5 | 72.1 | 74.7 |
| | Zhang et al. [72] | 54.9 | 58.0 | 66.2 | 71.1 | 72.8 |
| WiFi: 2D AoA | 3D-ID | 85.3 | **91.7** | **96.3** | **99.0** | **100** |

Table 1: Overall comparison of our system with existing approaches using ranking accuracy (measured by %).

Additionally, our system significantly outperforms the existing WiFi-based person identification systems. In particular, our system performs 22% to 32% better than that of the prior WiFi-based systems for rank-1 accuracy. For other accuracies, our system consistently performs better. This is because CSI is susceptible to background environment changes. Although the time-frequency feature of WiFi extracts human dynamics, it may vary with the WiFi device layouts/configurations and the walking activities of the person. It is thus very hard for existing WiFi-based systems to reliably distinguish different people across various environments without leveraging the intrinsic features of a person (e.g., static or dynamic biometrics). Instead, our system utilizes spatial information to generate the image of a person and further extracts the static and dynamic biometrics of the person, enabling a more robust and secure person Re-ID system.

### 6.3 Impact of Unseen Environments

The person Re-ID system could be trained in one environment and recognize a person in unseen environments, where the system has not been trained. Thus, we study the impact of unseen environments on the performance of our system and the traditional WiFi-based system [29]. In particular, for both systems, we collect WiFi data in three environments to train the model and test it in the unseen (remaining) environment. Figure 12 shows the CMC curves in four different environments for both the proposed 3D-ID system (i.e., solid lines) and the traditional WiFi-based (abbreviated as Trad WiFi) system (i.e., dotted lines). Compared with the overall performance evaluation, this is a more challenging task as the person needs to be recognized in brand new environments with totally different backgrounds, transceiver positions, and orientations.

As shown in Figure 12, we can observe that there is no significant performance difference for our system across different environments. Specifically, the rank-1 accuracies of all environments are more than 83% and rank-3 accuracies are about 96%. However, the traditional WiFi-based system only achieves the rank-1 accuracy of around 52% in unseen environments. The reason is that traditional WiFi-based systems are tied to the WiFi network configuration, user activities, and the environment. They are sensitive to activity variations and could not completely separate the signal reflections of humans from the surrounding environment. For example, the walking information derived in prior WiFi-based systems changes arbitrarily due to the changes in walking path and direction, and layout of transceivers. Therefore, they are not performing well when deployed in a new environment. The above result demonstrates that our system is capable of re-identifying a person even if the environment has never been seen before.

### 6.4 Performance Under Attacks

Next, we evaluate the system performance under attacks described in Section 4.2. To deceive the system, the adversary either changes clothes styles from skinny to baggy or from skirt to pants. In addition, the adversary changed the clothes to different colors. The results are shown in Figure 13. We can observe that rank-1 accuracy is 85.1% and there is almost no performance degradation. The reason is that WiFi signals can traverse clothes and reflect off the human body. Thus, unlike vision-based systems, our system is robust to appearance.

To evade detection and identification, the adversary hides and walks behind a wooden screen (i.e., NLoS). As shown in Figure 13, our system still achieves a rank-1 accuracy of 83.2%. Such a result shows that our system can identify people even under the NLoS scenario, where the vision-based systems fail completely. This is because the WiFi signals can penetrate obstacles, while visible light will be totally blocked.

With some knowledge of our Re-ID system, an adversary can conduct advanced attacks by deliberately changing walking behaviors to spoof the system. As illustrated in Figure 13, the rank-1 accuracy is reduced to 68.4%. The result shows that changing walking behaviors could degrade performance. However, as also shown in Figure 13, our system can still utilize body shape features to identify people (e.g., rank-1 accuracy is about 76%). This is because the body shape is persistent and cannot be easily changed.

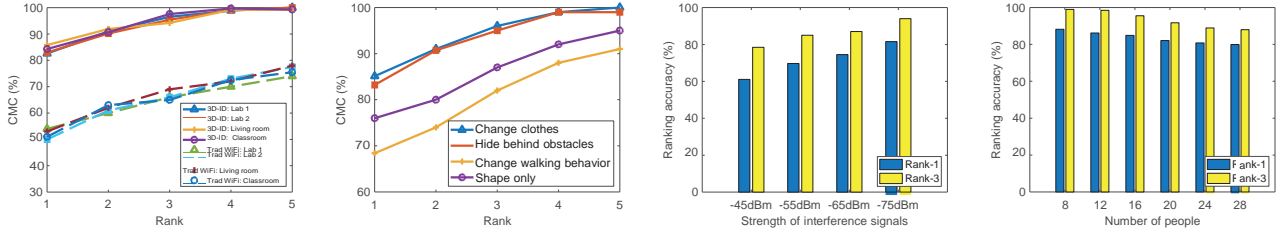

Figure 12: System performance comparison under unseen environments.

Figure 13: System performance under clothes change, NLoS, and behavior change.

Figure 14: System performance under wireless interference.

Figure 15: System performance under the different number of people.

Moreover, wireless interferences could have an impact on the WiFi-based person Re-ID system. Therefore, the adversary tries to attack our system using different strengths of interference signals at the same wireless channel that is used by WiFi devices. As shown in Figure 14, we report rank-1 and rank-3 accuracies under interference signals of -45dBm, -55dBm, -65dBm, and -75dBm respectively. We can observe that when the interference strength is reduced, the system performance becomes better. We observe that our system can work properly with the rank-3 accuracy of 79% under very strong wireless interferences (i.e., -45dBm). We acknowledge that jamming in WiFi networks may disrupt both sensing and communication as it overpowers the original signals. But such jamming signals are very obvious and can be easily detected and neutralized.

### 6.5 Impact of Number of People

Ranking accuracy could be affected when increasing the number of people. Therefore, we evaluate the system performance by varying the number of people in the test dataset. Figure 15 shows the rank-1 and rank-3 accuracies when the number of people is varied from 8 to 28. We observe that rank-1 and rank-3 accuracies are around 80% and 88% for 28 people. They are 88% and 99% for 8 people, respectively. We can observe that as we reduce the number of people, the ranking accuracy increases. This is because it is less likely to have two people with similar body shapes and walking patterns in a smaller group of people. In addition, our system still achieves a good rank-3 accuracy on a dataset that includes all the people.

### 6.6 Impact of Number of Antennas

Although new-generation WiFi devices could support up to 16 antennas, many home WiFi devices have a smaller number of antennas. Hence, we study the impact of the number of antennas by equipping each receiver with 9, 7, and 5 antennas, respectively. Figure 16 shows that the rank-1 accuracies for using 9, 7, and 5 antennas are around 85%, 83%, and 80%, respectively. The result demonstrates that when the number of antennas increases, the system performance becomes better. It is because more antennas lead to a higher resolution of 2D AoA and thus can visualize the subject more precisely. Moreover, our system can work properly with only 5 antennas which could be the default number of antennas on next-generation WiFi devices.

### 6.7 Impact of Walking Duration

In our evaluation, we utilize 100 consecutive frames (probe sequence length) by default for each person in the test. Note that our system generates 30 frames per second and 100 frames represent 3.3s. However, the duration of walking can vary from person to person. Therefore, we study the impact of walking duration with 50 frames (1.7s), 100 frames (3.3s), and 200 frames (6.7s). As shown in Figure 17, a longer duration can improve system performance as our system can obtain more temporal information. Still, the rank-1 accuracy is 82% even when we only use 50 frames. The reason is that our system identifies people leveraging both static and dynamic biometrics information. Even with limited dynamic walking information, our system can still utilize static shape information to maintain a good performance.

### 6.8 Impact of Sensing Range

To study the impact of sensing range change on performance, our system was evaluated with various sensing ranges: 2.8*m*, 3.5*m*, and 4.2*m*. According to the CMC curves shown in Figure 18, we can observe that the system performance increases as the sensing range decreases. This is due to the fact that a shorter propagation distance results in a stronger received signal. However, the system performance is still comparable at different distances. Thus, our system can work well in a typical room with a variety of distances. Note that by employing higher gain antennas, the WiFi transmission power can be increased and the corresponding maximum sensing range will be extended.

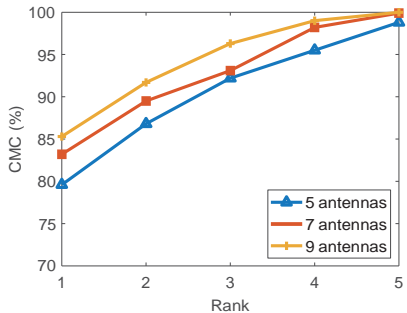 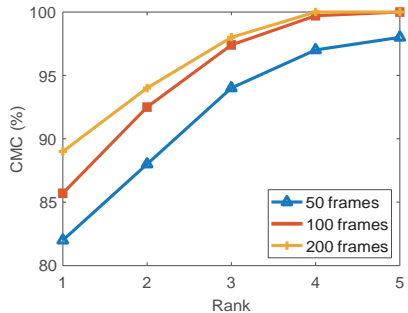 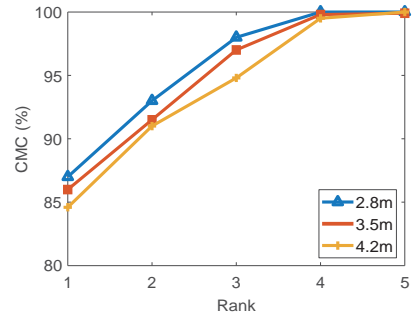

Figure 16: System performance under the different number of receiving antennas.

Figure 17: System performance under different walking durations.

Figure 18: System performance under different sensing ranges.

## 7 Discussion

The 3D-ID system shows that our proposed WiFi vision-based person Re-ID is very promising. However, the current implementation still has some limitations.

**Crowded Space.** Currently, we assume that only one person is walking in an indoor environment. Identifying a large number of people simultaneously in a crowded space (e.g., a train station and shopping mall) is still a challenge. Under crowded spaces, human bodies could overlap, and a large number of people would cause very complicated signal reflections. This is also a challenging problem in the computer vision community. One promising research direction to address this challenge is to segment each person based on high-resolution 2D AoA images and then perform identification one by one, similar to the approaches used in computer vision. Still, it is an open question of how many persons can be supported simultaneously with the proposed WiFi vision-based approach on next-generation WiFi devices.

**Sensing Range.** Although we have demonstrated our system can work at several meters' sensing range, it is still limited when compared to a camera-based system. As our approach relies on the signal reflections, whose power is normally several orders of magnitude weaker than that of the signal that went through LoS propagation. The sensing range of the proposed WiFi vision-based approach is thus much shorter than the communication range of the WiFi. This issue, however, could be mitigated by leveraging the pervasively deployed WiFi devices, i.e., every location could be covered by WiFi devices several meters away, or by leveraging directional antennas.

**Limited Users.** Our current evaluation only involves 28 participants, which is limited. Although the performance of our system is comparable to the computer vision-based approaches under the same number of participants, the system could benefit from a stress test by involving a large number of participants. After the COVID-19 pandemic, we would like to recruit more participants for a long-term study.

## 8 Conclusion

Person Re-ID in traditional optical camera-based systems is challenging due to changes in the appearance of people, occlusions, and unconstrained human poses. We propose a WiFi vision-based person Re-ID system, 3D-ID, which is very promising to mitigate these challenges and augment traditional camera-based systems. Specifically, we exploit multiple antennas on next-generation WiFi devices and 2D AoA of the WiFi signal reflections to visualize a person in the physical environment. Our system extracts intrinsic features of the body shape and dynamic walking patterns from the digitized 3D human body for person Re-ID. Our system is thus resistant to the changes in the appearance of people as well as the unconstrained poses. Our system can also work under NLoS scenarios as the WiFi signals traverse occlusions and actively illuminate the human body. Extensive experiments in various indoor environments demonstrate that the 3D-ID system is effective in identifying a number of people and that it can achieve an overall rank-1 accuracy of 85.3%. Our system is also resistant to various attacks.

## Acknowledgments

We thank the anonymous reviewers and our shepherd for their insightful feedback. This work was partially supported by the NSF Grants CNS-1910519, CNS-2131143, DGE-1565215, DGE-2146354, CNS-2120396, CCF-1909963, and CCF-2211163.